\documentclass[10pt,twocolumn,letterpaper]{article}

\usepackage{iccv}
\usepackage{times}
\usepackage{epsfig}
\usepackage{graphicx}
\usepackage{amsmath}
\usepackage{amssymb}

\usepackage[breaklinks=true,bookmarks=false]{hyperref}

\iccvfinalcopy 

\def\iccvPaperID{****} 

\ificcvfinal\fi

\begin{document}

\title{Task-Specific Data Augmentation and Inference Processing for VIPriors Instance Segmentation Challenge}

\author{Bo Yan, Xingran Zhao, Yadong Li, Hongbin Wang \\
\small
Ant Group, China \\
}

\maketitle
\ificcvfinal\thispagestyle{empty}\fi

\begin{abstract}

Instance segmentation is applied widely in image editing, image analysis and autonomous driving, etc. However, insufficient data is a common problem in practical applications. The Visual Inductive Priors(VIPriors) Instance Segmentation Challenge has focused on this problem. VIPriors for Data-Efficient Computer Vision Challenges ask competitors to train models from scratch in a data-deficient setting, but there are some visual inductive priors that can be used. 
In order to address the VIPriors instance segmentation problem, we designed a Task-Specific Data Augmentation(TS-DA) strategy and Inference Processing(TS-IP) strategy. The main purpose of task-specific data augmentation strategy is to tackle the data-deficient problem. And in order to make the most of visual inductive priors, we designed a task-specific inference processing strategy. We demonstrate the applicability of proposed method on VIPriors Instance Segmentation Challenge. The segmentation model applied is Hybrid Task Cascade based detector on the Swin-Base  based CBNetV2 backbone. Experimental results demonstrate that proposed method can achieve a competitive result on the test set of 2022 VIPriors Instance Segmentation Challenge, with 0.531 AP@0.50:0.95.

\end{abstract}

\section{Introduction}

Instance segmentation applies widely in image editing, image composition, autonomous driving, etc. Instance segmentation is a fundamental problem in computer vision. Deep learning-based methods have achieved promising results for image instance segmentation over the past few years, such as Mask R-CNN~\cite{he2017mask}, PANet~\cite{liu2018path}, TensorMask~\cite{chen2019tensormask},  CenterMask~\cite{wang2020centermask}. SOLO series~\cite{wang2020solo, wang2020solov2}. The main objective of the VIPriors Instance Segmentation Challenge is to tackle the segmentation of basketball and individual humans including players, coaches and referees on a basketball court. Different from previous studies, VIPriors Instance Segmentation Challenge does not allow using any pre-trained model, and training data are deficient.

\hfill

In order to address the problem of data-deficient, we designed a task-specific data augmentation(TS-DA) strategy, which mainly contains two components, specific copy-paste and base-transform strategies. The main purpose of TS-DA strategy is to increase the diversity of data distribution that can help with the data-deficient problem. In order to make the most of visual inductive priors, we designed a task-specific inference processing(TS-IP) strategy that mainly includes two parts, inference cropping and max score filtering strategies.

\hfill

\section{Methodology}

In this section, we first illustrate the task-specific data augmentation strategy in Sec.2.1. The task-specific inference processing strategy is presented in Sec.2.2. And the segmentation model and training strategy is introduced in Sec.2.3.

\subsection{Task-Specific Data Augmentation}

For VIPriors instance segmentation, although the scenario seems simple because the task is only to segment basketball and players, coaches and referees on images recorded of a basketball court, total images including train, validation and test are also insufficient. In order to generate enough image instances with wider distribution and make the model performs better and more robust, we designed a task-specific data augmentation strategy that mainly includes two parts, specific copy-paste and base-transform strategies.

Copy-Paste~\cite{9578639}, which copies objects from one image to another, is particularly useful for instance segmentation. Different from the original copy-paste augmentation, we designed a specific copy-paste strategy including view-specific copy-paste and ball-specific copy-paste. 

We first cropped all objects including persons and basketballs on training and validation datasets, then saved the cropped images and their corresponding mask annotations to storage for the next specific copy-paste strategy. 

We can know there are two views including the left view and right view for all images by the visual inductive priors. And we can get the view of image by the image name, if the prefix of image name end with ``0", the image comes from right view, otherwise left view. We designed a view-specific copy-paste strategy that copies and pastes the object into a specific location according to the view of image.

\begin{figure}[h]
    \centering
    \begin{minipage}[b]{0.5\textwidth}
    \includegraphics[width=1\textwidth]{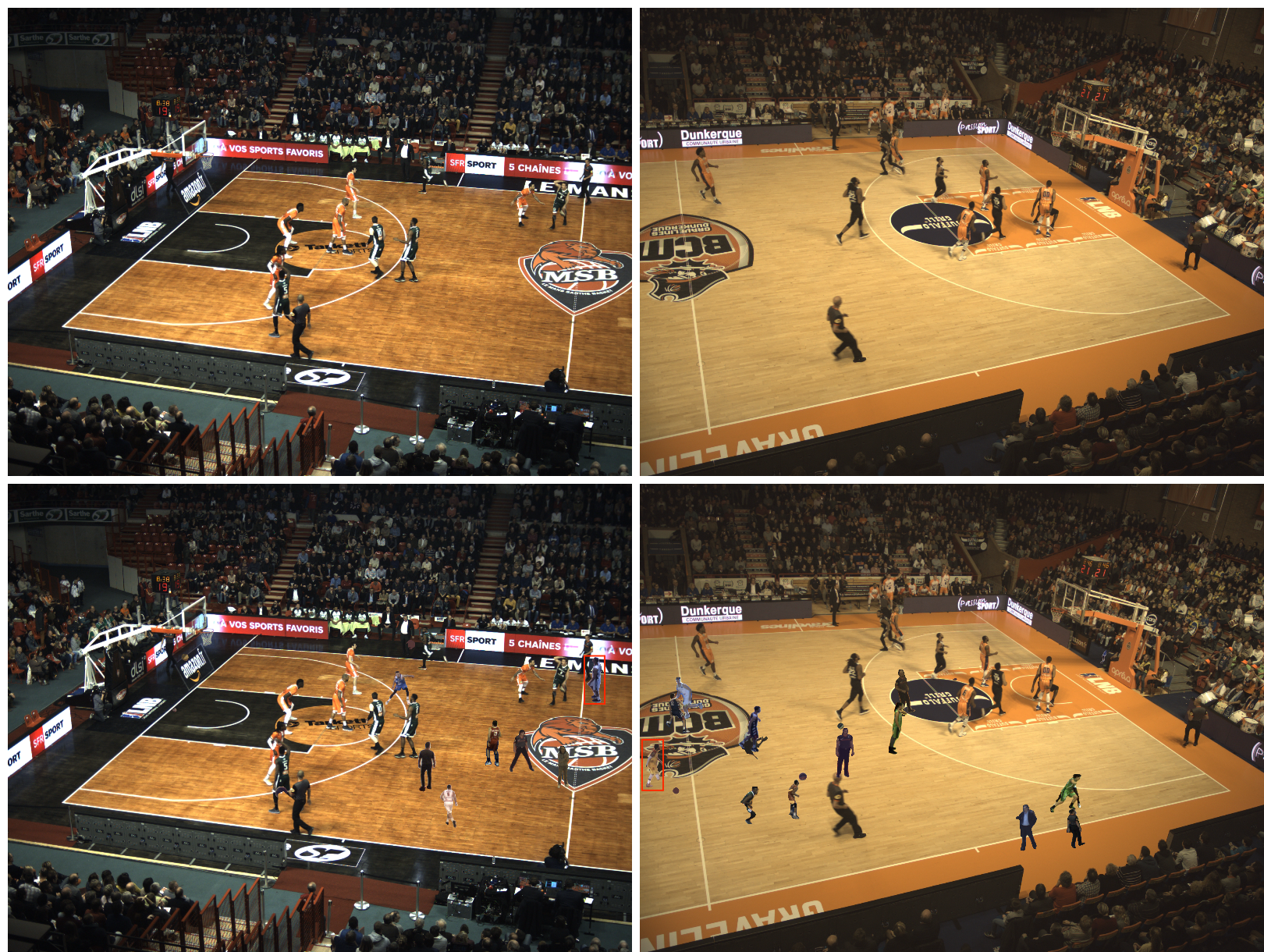} \\
    \centerline{(a) image from right view \hspace{2em} (b) image from left view \hspace{1em}}%
    \\
    \end{minipage}
\caption{Augmented image examples of view-specific copy-paste.} \label{fig1}
\end{figure}

For a given image, we would copy-paste several cropped persons and balls to it. If the image is come from right view, the copy-paste location constraints are Formula \ref{con1} and \ref{con3}, otherwise are Formula \ref{con2} and \ref{con3}. Our proposed view-specific copy-paste strategy can make the copy-paste objects more evenly distributed on the basketball court including the left and right borders, but excluding the auditorium. Figure \ref{fig1} shows examples of view-specific copy-paste strategy method on the training set, above are original images and below are corresponding augmented images.

\begin{equation}
\frac{w}{5} \leq x_{min} \leq w \label{con1}
\end{equation}
\begin{equation}
0 \leq x_{min} \leq w-\frac{w}{5} \label{con2}
\end{equation}
\begin{equation}
\frac{h}{2}-\frac{h}{5} \leq y_{min} \leq \frac{h}{2}+\frac{h}{5} \label{con3}
\end{equation}

Where $w$ and $h$ are the width and height of the given image, ($x_{min}$, $y_{min}$) denotes the top left coordinate of a person or ball that is being pasted.

\hfill

After the view-specific copy-paste phase, in order to further improve the segmentation performance of basketball, we designed a ball-specific copy-paste strategy including man-ball interaction copy-paste and pure-ball generation strategies.

The purpose of the man-ball interaction copy-paste strategy is to improve the segmentation performance of the ball when it interacts with person, For a given image, we will select several target persons on basketball court, and paste cropped balls around them, the copy-paste location constraints are as follows.

\begin{equation}
s_{x_{min}} \leq x_{min} \leq s_{x_{max}} \label{con4}
\end{equation}
\begin{equation}
s_{y_{min}} \leq y_{min} \leq s_{y_{max}} \label{con5}
\end{equation}

\begin{figure}[h]
    \centering
    \begin{minipage}[b]{0.5\textwidth}
    \includegraphics[width=1\textwidth]{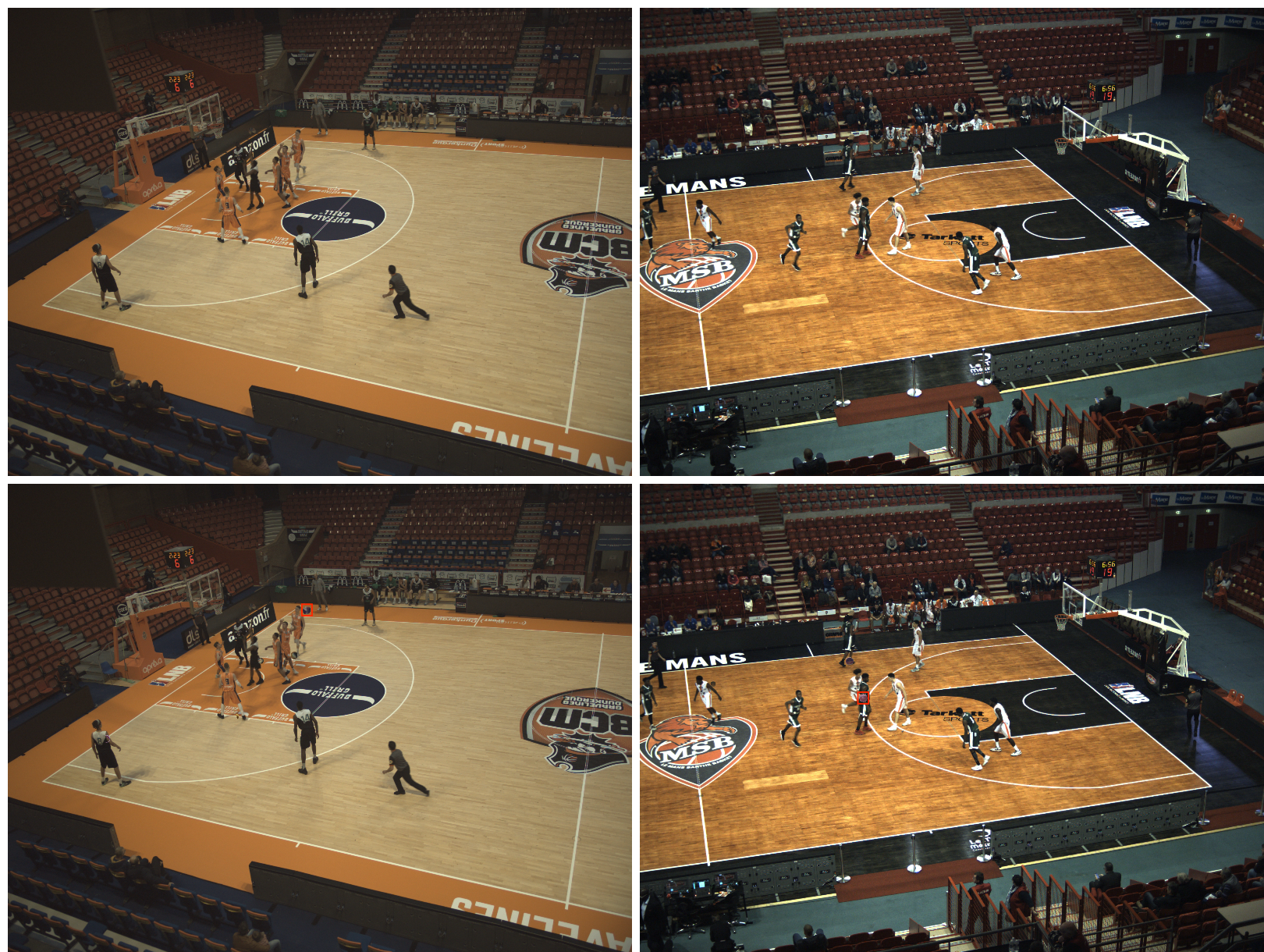} \\
    \end{minipage}
\caption{Augmented image examples of man-ball interaction copy-paste.} \label{fig2}
\end{figure}

Where ($s_{x_{min}}$, $s_{y_{min}}$) and ($s_{x_{max}}$, $s_{y_{max}}$) are the top left and  bottom right coordinate of selected person respectively, ($x_{min}$, $y_{min}$) denotes the top left coordinates of a ball that is being pasted. Figure \ref{fig2} shows examples of man-ball interaction copy-paste strategy method on the training set, above are original images and below are corresponding augmented images.

\hfill

\begin{figure}[h]
    \centering
    \begin{minipage}[b]{0.5\textwidth}
    \includegraphics[width=0.9\textwidth]{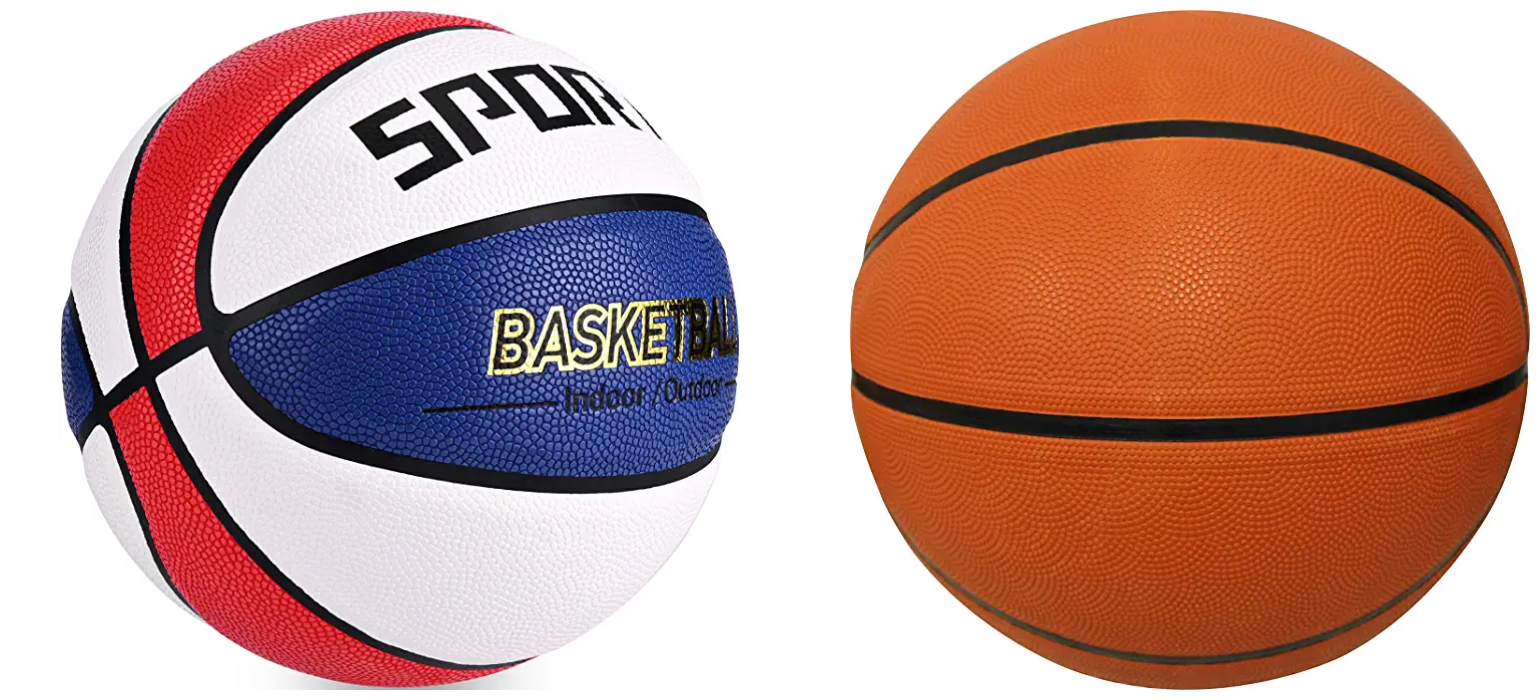} \\
    \centerline{(a) colorful ball \hspace{5em} (b) pure ball \hspace{3em}}%
    \\
    \end{minipage}
\caption{Image examples of common basketball types.} \label{fig3}
\end{figure}

It is widely known that there are mainly two types of basketball by the visual inductive priors, as shown in Figure \ref{fig3}. The basketball that likes shown in Figure \ref{fig3} (a) consists of at least two colors, we call it colorful ball, otherwise the ball mainly consists of single color, we call it pure ball. 

For a cropped ball image, we would directly change the RGB value of image but the mask is unchanged as an approximate method of generating pure ball. For example, if we want to get a brown pure ball, the Red value of image would be changed into a random value from 80 to 90, and the Green and Blue value would be changed into a random value from 50 to 60. The balls in the training and validation set are mainly colorful balls, so we use pure-ball generation method with probability to generate more pure balls when copy-pasting balls including the view-specific copy-paste and man-ball interaction copy-paste phases.

\hfill

\begin{figure}[h]
    \centering
    \begin{minipage}[b]{0.5\textwidth}
    \includegraphics[width=1\textwidth]{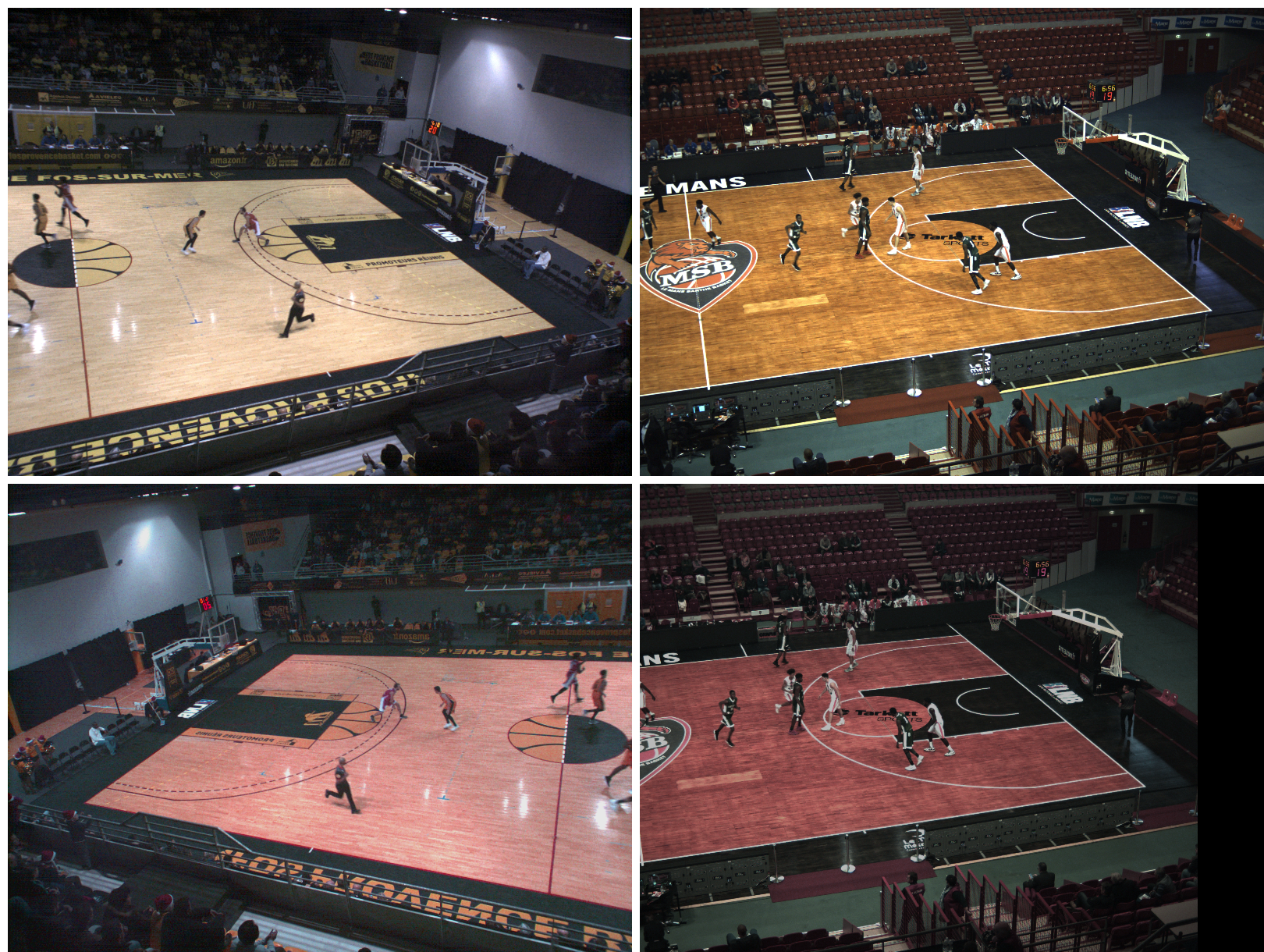} \\
    \end{minipage}
\caption{Augmented image examples of base-transform.} \label{fig4}
\end{figure}

After the specific copy-paste pipeline, the images are input to base-transform. The base-transform consists of two components, geometric transform, and photometric distortion transform. The geometric transform includes three detailed transforms, shear transform, rotate transform and translate transform, and we randomly choose one of them as geometric transform for every image. There are four detailed transforms for photometric distortion transform, random brightness, random contrast, random saturation and random hue, every image would be passed through these four transforms. After the base-transform pipeline, we can get an augmented image. As shown in Figure \ref{fig4}, above are the original images and below are examples of augmented results.

\hfill

To sum up, our task-specific data augmentation mainly consists of two components, specific copy-paste and base-transform strategies. The images are first input to a view-specific copy-paste pipeline, we copy-paste several persons and balls for every image with specific constraints. And then images are input to a  ball-specific copy-paste strategy, we copy-paste several balls around persons, which could help further improve the segmentation performance of basketball. We use the pure-ball generation method with probability to generate more pure balls when copy-pasting balls including the view-specific copy-paste and ball-specific copy-paste phases. Finally, the images are input to base-transform. Followed by the specific copy-paste and base-transform pipelines, we can get final augmented images as the input to the segmentation network.

\subsection{Task-Specific Inference Processing}

In order to make the most of visual inductive priors, we designed a task-specific inference processing strategy that mainly includes two parts, inference cropping and max score filtering strategies.

\begin{figure}[h]
    \centering
    \begin{minipage}[b]{0.5\textwidth}
    \includegraphics[width=1\textwidth]{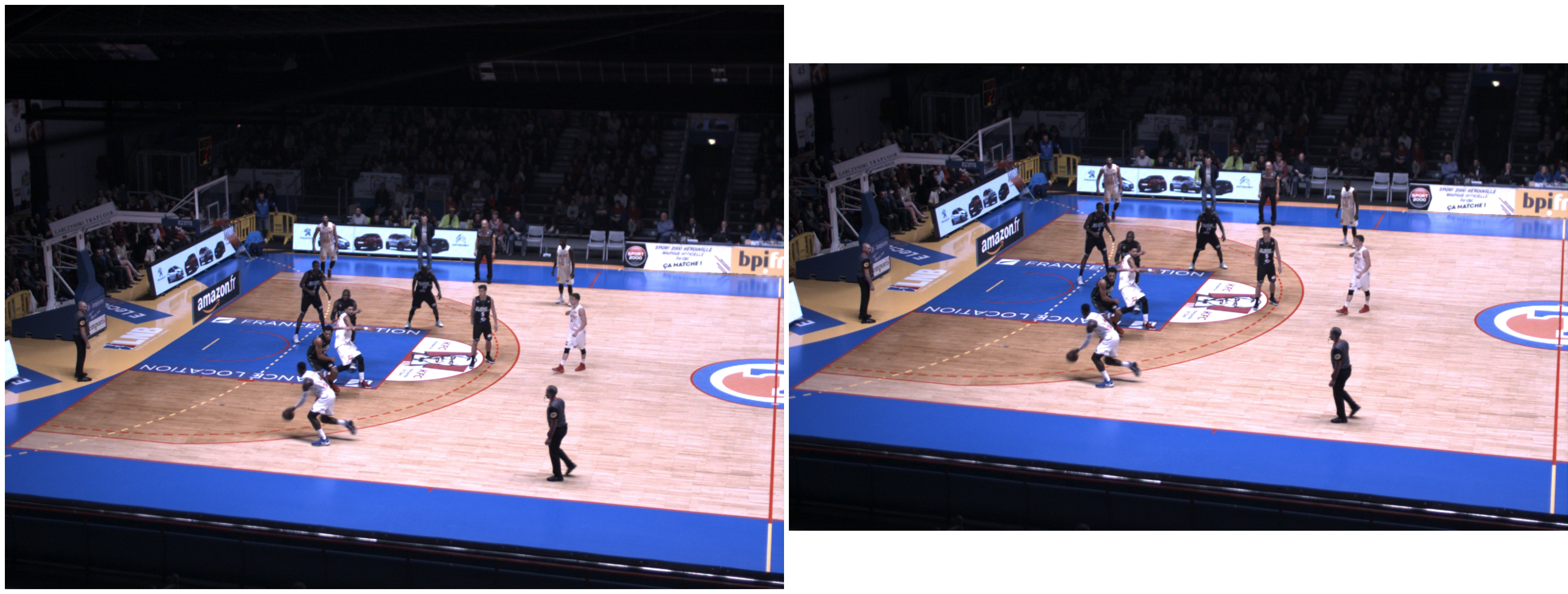} \\
    \end{minipage}
\caption{Image example of inference cropping on test set.} \label{fig5}
\end{figure}

Different from the common multi-scale training strategy, we resize the input images randomly from a small scale to large scale, then randomly cropped and padded them to a fixed scale, which is helpful to our inference cropping strategy. We can know that the upper part of image is mainly auditorium by visual inductive priors, so we cropped $1/5$ of image from above when inference, then input to the trained model and get the segmentation results, and invert the segmentation results to original image finally. Figure \ref{fig5} is an example of cropped image on the test set, left is the original image and right is the cropped image when inference. The inference cropping strategy can remove some invalid areas and improve model performance.

\hfill

\begin{figure}[h]
    \centering
    \begin{minipage}[b]{0.5\textwidth}
    \includegraphics[width=1\textwidth]{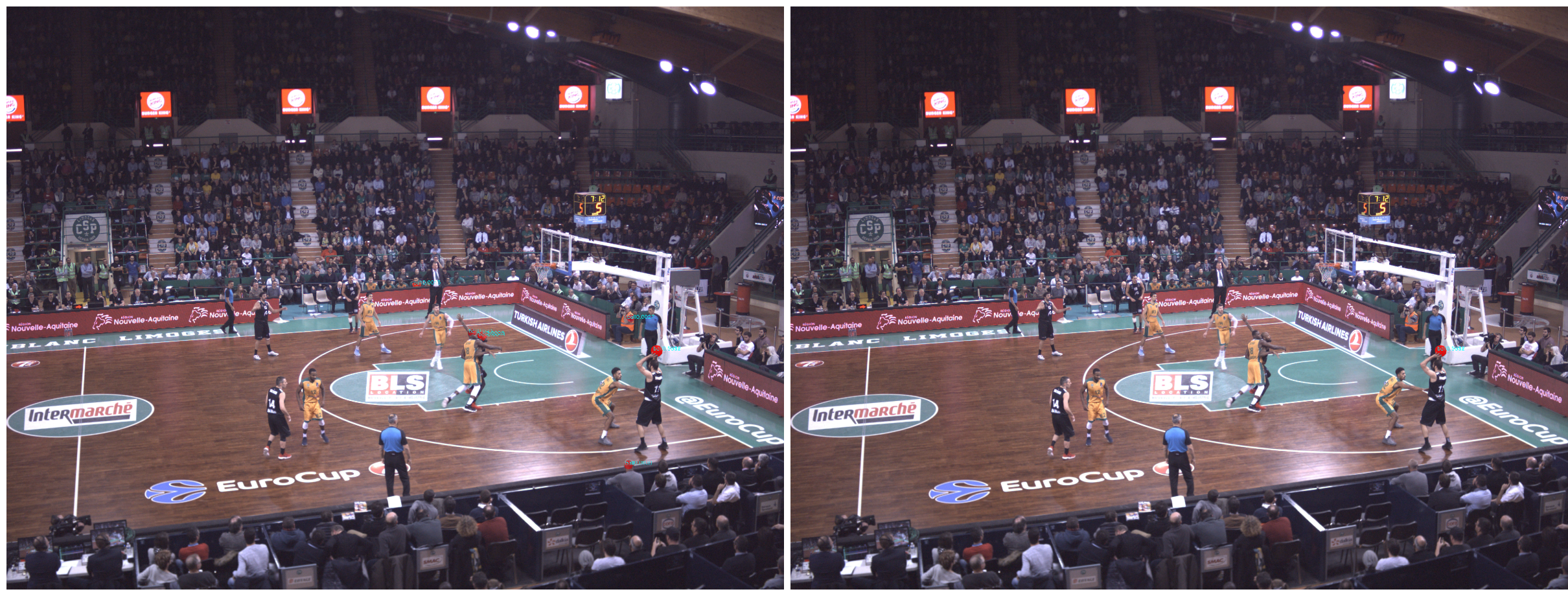} \\
    \end{minipage}
\caption{Result example of max score filtering on test set.} \label{fig6}
\end{figure}

We can know that a basketball court can only have one basketball at most by visual inductive priors, so we designed a max score filtering strategy for the segment results of basketball. For a given test image, we can get the segmentation results including human and basketball by the trained model, for the segment results of basketball, we sort the segmentation results by score and the maximum score result is preserved. And for the other segmentation results of ball, we will calculate the Intersection over Union(IoU) of bounding boxes and the bounding box of maximum score result, if IoU $ > 0$ the result will be preserved otherwise dropped. In addition, we also filtered out some balls with very small length and width($<10$) and very large length and width($>40$). Figure \ref{fig6} shows an example of max score filtering strategy on the test set, left is original result of model output and right is the result of max score filtering.

\hfill

\subsection{Segmentation Model and Training Strategy}

Our segmentation model is Hybrid Task Cascade(HTC)~\cite{chen2019hybrid} based detector on the CBSwin-Base backbone with CBFPN~\cite{Liang2021CBNetV2AC}. The Mask R-CNN with MaskIoU head named as Mask Scoring R-CNN~\cite{huang2019mask}, mask scoring can automatically learn the mask quality instead of relying on the classification confidence of the bounding box, so we add MaskIoU head to HTCMaskHead that can simply and effectively improve the performance of instance segmentation due to the alignment between mask quality and mask score. 

\begin{figure*}[ht]
	\centering
	\includegraphics[scale=0.5]{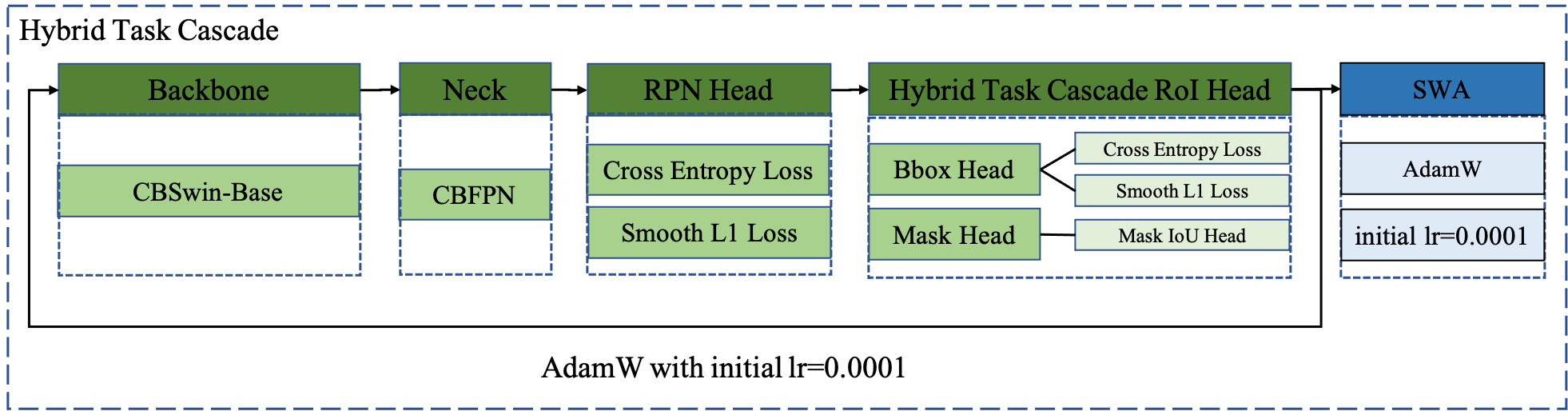}
	\caption{Key components of model architecture and training pipeline for VIPriors challenge.}
	\label{fig7}
\end{figure*}

We first train the model normally with proposed task-specific data augmentation strategy. When the model is converged, we use SWA~\cite{zhang2020swa} training strategy to finetune the model, which can make the model better and more robust. The key components of model architecture and training pipeline are shown in Figure \ref{fig7}.

\hfill

\section{Experiments}

\subsection{Training Details}

We train and evaluate our model from scratch without any pre-trained model and backbone. And we only use the training and validation datasets supplied by VIPriors Instance Segmentation Challenge to train the model. Firstly, we duplicated training and validation images 10 times without any change. The total images for training are 2460 including 1840 training images and 620 validation images. Then the images are input to our proposed task-specific data augmentation pipelines. And then the augmented images are randomly scaled from 820 to 3080 on the short side and up to 3680 on the long side, and after that randomly cropped and padded to (1920, 1440). Finally, the images are input to segmentation model.

Firstly, we train the model with Adam with decoupled weight decay(AdamW)~\cite{loshchilov2018decoupled} and initial learning rate(lr)=0.0001. When the model is converged, we use SWA training strategy to finetune the model, and the optimizer is also AdamW with initial lr=0.0001. 

\subsection{Experimental Results}

As shown in Table \ref{tab1}, our proposed method finally achieves 0.531 AP@0.50:0.95 on the test set of VIPriors Instance Segmentation Challenge.

\begin{table*}[h]
\begin{center}
\resizebox{.95\textwidth}{!}{
\begin{tabular}{lcccc}
\hline
Models & Training Data & Data Augmentation & Methods & mAP-Test\\
\hline
HTC-CBSwinBase & Train+Val & - & - & 0.396 \\
HTC-CBSwinBase+MaskIoU & Train+Val & - & - & 0.415 \\
HTC-CBSwinBase+MaskIoU & Train+Val & TS-DA & - & 0.472 \\
HTC-CBSwinBase+MaskIoU & Train+Val & TS-DA & +TTA & 0.477 \\
HTC-CBSwinBase+MaskIoU & Train+Val & TS-DA & +Mask Threshold Binary=0.4 & 0.489 \\
HTC-CBSwinBase+MaskIoU & Train+Val & TS-DA & +SWA & 0.512 \\
HTC-CBSwinBase+MaskIoU & Train+Val & TS-DA & + Inference Cropping & 0.522 \\
HTC-CBSwinBase+MaskIoU & Train+Val & TS-DA & + Max Score Filtering & 0.529 \\
HTC-CBSwinBase+MaskIoU & Train+Val & TS-DA & +Ensemble(Mask Loss Weight=2.0) & 0.531 \\
\hline
\end{tabular}
}
\end{center}

\caption{Experimental results on dataset of VIPriors challenge. ``+" indicates that it is added on the basis of previous method. \label{tab1}}
\end{table*}
\hfill

\subsection{Ablation Study}

This section elaborates on how we achieve the final result by ablation study to explain our method. 

Our baseline is HTC-CBSwinBase and soft nms~\cite{bodla2017soft} is used at the test stage for all experiments. The baseline achieves 0.396 mAP on test set. We add MaskIoU head to HTC-CBSwinBase, it can improve the baseline by 0.019 mAP. 

The experimental results show that TS-DA is a very effective strategy for this task, it can bring an improvement of 0.057 mAP. Then test time augmentation(TTA) is applied, our TTA strategy is flip horizontal and multi-scale test with scale factors (1.0, 1.5, 2.0), and TTA can bring a little improvement of 0.005 mAP. Then we attempt to adjust the parameters of test configs, we adjust the threshold of rcnn mask binary from 0.5 to 0.4, which can bring an improvement of 0.012 mAP unexpectedly. Then we use SWA training strategy to finetune the model and it can improve 0.023 mAP. And then our proposed inference cropping strategy can bring an improvement of 0.010 mAP. Then we use the proposed max score filtering strategy to filter the results of inference cropping strategy, it can improve 0.007 mAP and achieves 0.529 mAP on the test set. 

Finally, we train the model whose settings are the same as before but the loss weight of mask head is changed from 1.0 to 2.0, and then we ensemble the results of these two models, it achieves 0.531 mAP on the test set.

\hfill

\section{Conclusion}

In this paper, we introduce the proposed method to address the VIPriors instance segmentation problem. In order to tackle the data-deficient problem, we designed a Task-Specific Data Augmentation strategy. And also we designed a Task-Specific Inference Processing strategy according to the visual inductive priors. We demonstrate the applicability of proposed method on the VIPriors Instance Segmentation Challenge. Experimental results demonstrate that the proposed method can achieve a competitive result on the test set without external image or video data and pre-trained weights. Finally, our method achieves 0.531 AP@0.50:0.95 on the test set of VIPriors Instance Segmentation Challenge.


{\small
\bibliographystyle{ieee_fullname}
\bibliography{egbib}
}

\end{document}